\documentclass[10pt,twocolumn,letterpaper]{article}

\makeatletter
\@namedef{ver@everyshi.sty}{}
\makeatother

\usepackage[pagenumbers]{cvpr} 

\usepackage{graphicx}
\usepackage{amsmath}
\usepackage{amssymb}
\usepackage{booktabs}

\usepackage{adjustbox}
\usepackage{array}
\usepackage{makecell}
\usepackage{multirow}
\usepackage{dsfont}
\usepackage{diagbox}
\usepackage{pgf}
\usepackage{amsfonts}
\usepackage{pifont}
\usepackage[figuresright]{rotating}
\usepackage{subcaption}
\usepackage{stackengine}
\usepackage{booktabs, arydshln}
\usepackage[toc,page]{appendix}

\usepackage{multirow}

\newcolumntype{R}[2]{%
    >{\adjustbox{angle=#1,lap=\width-(#2)}\bgroup}%
    l%
    <{\egroup}%
}

\newcommand*\rot{\rotatebox{90}}

\makeatletter
\def\adl@drawiv#1#2#3{%
        \hskip.5\tabcolsep
        \xleaders#3{#2.5\@tempdimb #1{1}#2.5\@tempdimb}%
                #2\z@ plus1fil minus1fil\relax
        \hskip.5\tabcolsep}
\newcommand{\cdashlinelr}[1]{%
  \noalign{\vskip\aboverulesep
           \global\let\@dashdrawstore\adl@draw
           \global\let\adl@draw\adl@drawiv}
  \cdashline{#1}
  \noalign{\global\let\adl@draw\@dashdrawstore
           \vskip\belowrulesep}}
\makeatother

\usepackage[pagebackref,breaklinks,colorlinks]{hyperref}

\usepackage[capitalize]{cleveref}
\crefname{section}{Sec.}{Secs.}
\Crefname{section}{Section}{Sections}
\Crefname{table}{Table}{Tables}
\crefname{table}{Tab.}{Tabs.}

\begin{document}

\title{LegoFormer: Transformers for Block-by-Block Multi-view 3D Reconstruction}

\author{
\hspace{0.7cm}
Farid Yagubbayli$^1$
\hspace{1.45cm}
Yida Wang$^1$
\hspace{1.2cm}
Alessio Tonioni$^2$
\hspace{0.85cm}
Federico Tombari$^{1,2}$
\\
{
    \tt\small 
    farid.yagubbayli@tum.de
    \hspace{0.2cm}
    yida.wang@tum.de
    \hspace{0.2cm}
    alessiot@google.com
    \hspace{0.2cm}
    tombari@in.tum.de
}
\vspace{0.4cm}
\\
\multicolumn{1}{c}{
$^1$Technical University of Munich~~~
$^2$Google Inc.~~~
}
}

\maketitle

\begin{abstract}

Most modern deep learning-based multi-view 3D reconstruction techniques use RNNs or fusion modules to combine information from multiple images after independently encoding them. These two separate steps have loose connections and do not allow easy information sharing among views. We propose LegoFormer, a transformer model for voxel-based 3D reconstruction that uses the attention layers to share information among views during all computational stages. Moreover, instead of predicting each voxel independently, we propose to parametrize the output with a series of low-rank decomposition factors. This reformulation allows the prediction of an object as a set of independent regular structures then aggregated to obtain the final reconstruction. Experiments conducted on ShapeNet demonstrate the competitive performance of our model with respect to the state of the art while having increased interpretability thanks to the self-attention layers. We also show promising generalization results to real data.
\end{abstract}



\section{Introduction}

Efficient, accurate, and interaction-less multi-view 3D reconstruction methods are a prerequisite for robotic perception, 3D object modeling and augmented reality applications.
Structure-from-Motion (SfM) \cite{ozyesil2017survey} and Simultaneous Localization and Mapping (SLAM) \cite{fuentes2015visual} provide viable solutions, however, they either fail or generate only partial shapes when only a set of non-overlapping views are available due to the inability to find correspondences among them.
At the same time, this frequently occurs for practical applications: \eg, in 3D object modeling for online shopping only a small number of images from uncalibrated cameras are usually available. 

Recently, several CNN-based techniques were proposed to overcome these limitations and reconstruct objects as a voxel grid from a few uncalibrated images with limited coverage. 
Methods like 3D-R2N2 \cite{choy20163d} and LSM \cite{kar2017learning} treat multiple views as a sequence and utilize modified RNN models to fuse information from different views. 
RNNs, however, comes with drawbacks, such as long processing times, the difficulty of processing longer input sequences, and not being permutation invariant. 
More recent work \cite{xie2019pix2vox, xie2020pix2vox++} proposed to replace the RNN model by first decoding each view separately into a full volume and then fusing them with independent fusion and refinement modules in the network.
The modules need, however, to be trained progressively \cite{xie2019pix2vox}, making these solutions more complex to replicate. 
Lastly, all prior work predicts independently every voxel in the output 3D volume which can lead to non-regular reconstructions unless structural properties like smoothness and continuity are enforced explicitly.

\begin{figure}
    \centering
    \includegraphics[width=\columnwidth]{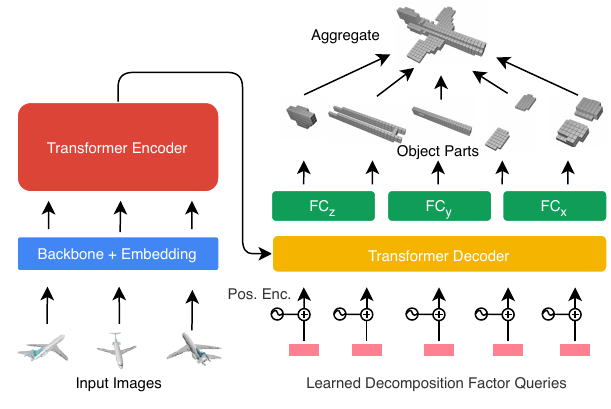}
    \caption{Given multiple views of an object, LegoFormer-M predicts a full 3D model as an aggregation of independently predicted \textit{parts}. The subdivision of the 3D model into parts is obtained in a self-supervised way thanks to our problem formulation.}
    \label{fig:architecture}
\end{figure}

To address these drawbacks, we propose LegoFormer: a model that uses a transformer network \cite{vaswani2017attention} to jointly encode all views and predict a 3D reconstruction in the form of its tensor decomposition. 
\cref{fig:architecture} illustrates a high-level overview of LegoFormer.
We argue that transformers are particularly suited to the task at hand since their attention mechanism allows information sharing across the inputs token, i.e., the views in our case. 
This is in contrast with previous work based on CNNs and RNNs, where the knowledge from all views becomes available only once the encoding phase ends. 
For the special case of only one input view available, the attention mechanism would not provide any advantage.
Therefore, we propose a special model for this task where patches from the same image are used as independent encoder inputs similarly to what is done in Vision Transformer (ViT)  \cite{dosovitskiy2020image}. 
To differentiate between the two cases, the models are respectively named LegoFormer-M (multi-view) and LegoFormer-S (single-view).

For both models, we observe that most of the objects we want to reconstruct are man-made and therefore expected to have a certain regular structure.
To leverage this property, we propose to use tensor decomposition to efficiently represent a voxel volume $V \in \mathds{R}^{N \times N \times N}$ by $k$ decomposition factors of size $\pmb{d} \in \mathds{R}^{3 \times N}$. 
The decoder part of our architecture takes a set of learned encoded queries as input and transforms each one into three vectors (the decomposition factors). 
Taking a cross-product between each triplet of vectors gives a rank-1 estimation of the full output voxel grid.
We show how different queries generate different estimations corresponding to individual parts of the object, e.g., the body of a plane or the wings in \cref{fig:architecture}; combining the parts together provides then the full model.
Remarkably, we do not provide explicit supervision for the different parts but let the network figure out a good subdivision as it is the only way of minimizing the loss functions given the constraints on the output space. 

We evaluate our proposal on ShapeNet\cite{chang2015shapenet}, showing competitive performance against prior techniques with more regular predictions and improved interpretability.
We also test the generalization of our method on real images from Pix3D \cite{sun2018pix3d}.
To summarize our contributions: \newline
\indent i) we introduce LegoFormer, a novel architecture for single- and multi-view object reconstruction that uses self-attention for both jointly encoding the input views and predicting the output volume; \newline
\indent ii) we propose to parametrize the output grid by its decomposition factors to take advantage of the inherent structures in common objects, and reformulate the task as the prediction of these factors; \newline
\indent iii) we show how LegoFormer learns to decompose a shape into a sum of tensor decomposition factors without explicit supervision for it.
%

\section{Previous work}

\textbf{3D Reconstruction:}
Classic techniques like SfM \cite{ozyesil2017survey} and SLAM \cite{fuentes2015visual} rely on feature extraction and matching across different views. Both steps assume highly overlapping views and Lambertian surfaces that limit their use. With the availability of large 3D object datasets and advances in deep learning, a new category of learned methods emerged. The majority of techniques for multi-view 3D reconstruction \cite{choy20163d, kar2017learning, xie2019pix2vox, xie2020pix2vox++} use an encoder-decoder architecture to map 2D images into 3D volumes without using 3D surfaces as intermediate representation. The mapping is done individually on each image therefore discarding information shared among views. Methods like 3D-R2N2 \cite{choy20163d} and LSM \cite{kar2017learning} introduce recurrent models to combine information from individual 3D volumes into a single volume, while Pix2Vox \cite{xie2019pix2vox, xie2020pix2vox++} uses a dedicated fusion unit. \cite{spezialetti2020divide} regresses the view poses and uses silhouettes to build an initial volume, then, as also done in LSM and Pix2Vox, uses a refiner unit that makes corrections to the predicted volumes based on the shape priors acquired during training. In contrast to these works, LegoFormer combines encoding, fusion, and decoding under a single transformer model. This structure allows the use of information from all the views in every stage of the process, resulting in a more tightly integrated framework.

\textbf{Transformers:}
Transformers have shown enormous success in a large variety of NLP tasks since their introduction by \cite{vaswani2017attention}. Recent work started to utilize transformer models for computer vision tasks by reformulating the inputs, which are mostly 2D RGB images, as a sequence of tokens \cite{dosovitskiy2020image, carion2020end, bertasius2021space}. ViT \cite{dosovitskiy2020image} is one of the earliest image classification methods that use a transformer encoder directly on a sequence of image patches. DETR \cite{carion2020end} combines a pre-trained CNN backbone with an encoder-decoder transformer and predicts a sequence of bounding boxes in parallel. TimeSformer \cite{bertasius2021space} has extended ViT with a spatiotemporal attention mechanism and achieves higher performance with less training time in video understanding tasks. More recently, hybrid approaches like LeViT \cite{graham2021levit} emerged where the convolutional blocks are integrated into the transformer model. Concurrently to our work \cite{wang2021multi} explored the use of a transformer-based model for 3D object reconstruction. Although there are similarities between the two works, our decoding scheme is significantly different.
To the best of our knowledge, together with \cite{wang2021multi} we are the first works to explore the use of transformer-based models for 3D shape reconstruction from multiple-views.

\textbf{Compressed Representation:}
An efficient representation is crucial when working with detailed 3D objects. Techniques like Distance Field Compression \cite{jones2004distance}, Octree \cite{zeng2013octree} and Voxel Hashing \cite{niessner2013real} exploit sparseness of the voxel volume and achieve 5-10 times compression rates \cite{boyko2020tt}. These methods use specialized data structures that often require complex mapping to the original 3D volume. On the other hand, TT-TSDF \cite{boyko2020tt} uses Low-Rank Tensor Train decomposition to represent the 3D tensor with lower rank tensors and demonstrates its application on the TSDF volumes. PARAFAC \cite{harshman1970foundations} and Tucker \cite{levin1965three, tucker1966some} are alternative general purpose decomposition algorithms where the latter has higher stability \cite{boyko2020tt}. Modified algorithms for boolean matrix decomposition were proposed by \cite{rukat2018tensormachine} and \cite{wan2020geometric} by introducing constraints over the decomposition factors. Matrix factorization is also used widely in recommendation systems to understand relationships \cite{koren2009matrix}. The LegoFormer output representation is similar to the one in  PARAFAC with a modification that bounds the decomposition factors in the range $[0, 1]$ to ensure binary decomposition. Negative numbers are not included in the prediction range to avoid "deletion" of the object parts during the aggregation phase.

\section{Methodology}

\subsection{LegoFormer}

\textbf{Overview:}
\cref{fig:architecture} provides an overview of the proposed architecture.
LegoFormer uses a vanilla pre-norm encoder/decoder transformer model \cite{nguyen2019transformers} to reconstruct a 3D occupancy grid of size $32^3$ from a set of 2D views. Each view is first mapped to an input token by a CNN-backbone. On the encoder side, the inputs are jointly encoded using the self-attention mechanism. Then the encoded inputs are passed to a non-autoregressive decoder together with a set of learned input queries. The decoder processes each query into a $32^3$ volume containing a "part" of the full object. Each "part" volume is parametrized by three vectors and calculated as the outer product between them. The final occupancy grid is the sum of the predicted parts. We do not supervise the subdivision of the objects into parts during training, which emerges naturally, instead, thanks to the architecture and the constrained output space.
Considering a single-view reconstruction, the attention mechanism on the encoder side would not do anything with the formulation described so far. For this reason, we also propose a variant of our method tailored for the single view case, where the input tokens are obtained from the patches of a single view instead of the full image. We will refer to the two variants of our method as LegoFormer-M and LegoFormer-S for the multi- and single-view cases to differentiate between the two scenarios, respectively.

\textbf{Backbone and Embedding:}
Due to the high spatial dimensionality of the input 2D images ($224 \times 224 \times 3$),  it is not feasible to feed them directly into the transformer, which typically has a much lower dimensionality. We use a VGG16 \cite{simonyan2014very}, pre-trained on ImageNet \cite{deng2009imagenet}, followed by $c$ additional  convolutional layers to map images to a compact feature representation $\psi$, as in \cite{carion2020end, xie2020pix2vox++}. Each convolutional layer is followed by batch normalization and a ReLU activation. The VGG16 is frozen during training while the convolutional layers are trained.

\emph{LegoFormer-M}: uses $c=3$ convolution blocks with a MaxPool after the second block with kernel size 2. The output features have size $\psi=8 \times 8 \times 64$ and are flattened and then projected using a single fully connected layer to obtain the input tokens for the transformer model.

\emph{LegoFormer-S}: uses $c=1$ convolution block that results in an output size $\psi=28 \times 28 \times 48$. The input sequence for LegoFormer-S is defined as a set of square patches of size $p=4$ extracted from $\psi$  resulting in $\frac{28}{4} \cdot \frac{28}{4} = 49$ input tokens. Each patch is projected using a single fully connected layer to obtain the input tokens for the transformer.

\textbf{Encoding:}
We use a vanilla transformer encoder with pre-norm residual connections \cite{nguyen2019transformers} to encode the input tokens. Each input token is allowed to attend to any other input while being encoded, so no self-attention mask is used. The pre-norm residual connections are preferred over post-norm connections due to the increased training stability. We do not use any positional encoding in LegoFormer-M to achieve input permutation invariance. Nevertheless, in LegoFormer-S we add to the input tokens a 2D positional sin-cos encoding \cite{parmar2018image} to inject the spatial relation between patches.

\textbf{Decoding:}
The decoder also uses the vanilla pre-norm transformer \cite{nguyen2019transformers} and a non-autoregressive formulation where the predictions are made in parallel. Following DETR \cite{carion2020end} we use learned decoder inputs, denoted decomposition factor queries, as input at the first decoder layer. The queries of the same size as the transformer dimensionality are initialized from a normal distribution with $\mu=0$ and $\sigma=1$. In contrast to the encoder, an attention mask is applied to prevent queries from attending to themselves. We experimentally found that such masking leads to a slight increase in performance compared to no masking. A 1D positional sin-cos encoding \cite{parmar2018image} is added to each learned query before passing it to the decoder. This addition is necessary to help distinguish between queries and avoid the collapse of all outputs to the same value during training.

\textbf{Output:}
The raw decoder outputs $Y_i$ are linearly projected using three fully connected layers $FC_{\{z, y, x\}}$ into three vectors of size $\mathds{R}^{32}$ corresponding to the decomposition factor components. After applying a sigmoid activation $\sigma$, we use a cross-product ($\otimes$) between the vectors to obtain a rank-1 estimation of the entire voxel grid. Each query is decoded into a rank-1 estimation, and all of them are combined by sum-aggregation to obtain the final reconstruction $P$. Formally, this is defined as

\begin{align}
    &\pmb{z}_i, \pmb{y}_i, \pmb{x}_i = FC_3(Y_i)
    \\
    &\pmb{z}_i, \pmb{y}_i, \pmb{x}_i = \sigma(\pmb{z}_i), \sigma(\pmb{y}_i), \sigma(\pmb{x}_i)
    \\
    &P = \text{min} (1, \sum_i \pmb{z}_i \otimes \pmb{y}_i \otimes \pmb{x}_i)
\end{align}

As the target values are in the range $[0, 1]$ the sigmoid is a suitable activation function. We clip the aggregated volume to prevent voxels with values larger than 1 after the sum-aggregation. The clipping also replicates the rules of boolean algebra where $1+1=1$ \cite{miettinen2020recent}

\textbf{Loss function:}
We use Mean-Squared-Error as the loss function to train the network. It is calculated between the predicted reconstruction and the ground-truth volume $G$ as follows:

\begin{equation}
    \mathcal{L}_{mse}(P, G) = \frac{1}{N^3} \sum_{i=0}^N \sum_{j=0}^N \sum_{k=0}^N (P_{i, j, k} - G_{i, j, k})^2
\end{equation}

\subsection{Alternative Schemes}
\label{sec:alternative_schemes}
During the development of this work we explored several alternatives to the tensor decomposition-based schema for the decoder part of our network. The alternative approaches subdivide the output volume into fixed patches and predict each one explicitly. Three such schemes are described below. Note that they are not strictly related to the proposed LegoFormer but they might be relevant for practitioners in the field.  We will show a comparison between all alternatives in  \cref{sec:other_schemas_experimental}.

\textbf{Naive}:  A simple approach for the output sequence subdivides the output volume into an ordered list of 3D patches of size $4^3$ and predicts them sequentially using a transformer decoder in an auto-regressive fashion. A causal mask \cite{vaswani2017attention} for the decoder-side attention is used during the training to prevent the decoder inputs from attending to future elements of the sequence. Formally, 

\begin{align}
    &Y_i = \mathcal{D}(\mathcal{E}, P_{0 \dots i-1})
    \\
    &P_i = \sigma(FC(Y_i))
    \\
    &P = \text{reshape}(P_0, P_1, \dots, P_{n})
\end{align}

Where $\mathcal{D}$ and $\mathcal{E}$ are the transformer decoder and the encoded views, respectively. Afterward, the predicted 3D patches are stitched together to obtain the total volume.

\textbf{Naive-nAR}: The above approach will result in low throughput performance during the inference due to the patches being predicted one by one. An alternative way, denoted "Naive-nAR", is predicting all patches simultaneously using a non-autoregressive decoder, which will significantly decrease the inference time. Formally, 

\begin{align}
    &P_i = \sigma(FC(\mathcal{D}(\mathcal{E}, \pmb{l}_i)))
    \\
    &P = \text{reshape}(P_0, P_1, \dots, P_{n})
\end{align}

where $\pmb{l}_i$ is a learned query as in the LegoFormer.

\textbf{Naive-Full}: Predicting the output volume in patches is computationally expensive and not scalable. Given a grid of side $32$ and a patch of side $4$, it will take $8^3=512$ predictions to cover the whole volume. Although transformers can efficiently attend to long sequences, the memory requirement grows quadratically \cite{ainslie2020etc} with the output sequence length. In order to reduce the number of predictions required, the whole volume can be predicted at once using a single learned query. However, this sacrifices the use of attention on the decoder. The scheme can be defined as,
\begin{align}
    &P = \sigma(FC(\mathcal{D}(\mathcal{E}, \pmb{l}))
\end{align}

\setlength{\tabcolsep}{5pt} 
\renewcommand{\arraystretch}{1} 

\begin{table*}[ht]
  \centering
\scalebox{0.8}{
\begin{tabular}{l|ccccccccc}
            \toprule
            & \multicolumn{9}{c}{Number of Views} \\
            \cmidrule(r){2-10}
            Model
            & 1 & 2 & 3 & 4 
            & 5 & 8 & 12 & 16 & 20 \\
            \midrule
\multicolumn{2}{l}{\textbf{Metric: IoU}} \\
\midrule
3D-R2N2\cite{choy20163d}      
    & 0.560   & 0.603   & 0.617   & 0.625    
    & 0.634     & 0.635     & 0.636       & 0.636   &   0.636
    \\
AttSets \cite{yang2020robust}
    & 0.642  & 0.662    & 0.67    & 0.675
    & 0.677      & 0.685    & 0.688       & 0.692   &   0.693
    \\
Pix2Vox/F \cite{xie2019pix2vox}   
    & 0.634  & 0.660    & 0.668   & 0.673    
    & 0.676      & 0.680    & 0.682       & 0.684   &   0.684
    \\
Pix2Vox/A \cite{xie2019pix2vox}
    & 0.661  & 0.686    & 0.693   & 0.697    
    & 0.699      & 0.702    & 0.704       & 0.705   &   0.706
    \\
Pix2Vox++/F \cite{xie2020pix2vox++}  
    & 0.645  & 0.669    & 0.678   & 0.682    
    & 0.685      & 0.690    & 0.692       & 0.693   &   0.694
    \\
Pix2Vox++/A \cite{xie2020pix2vox++}
    & \textbf{0.670}  & \textbf{0.695}   & \textbf{0.704}   & \textbf{0.708}
    & \textbf{0.711}     & \textbf{0.715}    & \textbf{0.717}  & 0.718   &   0.719
    \\
LegoFormer-M 
    & 0.519  & 0.644    & 0.679   & 0.694    
    & 0.703      & 0.713    & \textbf{0.717}       & \textbf{0.719}   &   \textbf{0.721}
    \\
LegoFormer-S 
    & 0.655  & -   & -   & -    
    & -      & -     & -   & - & -
    \\
\midrule
\multicolumn{2}{l}{\textbf{Metric: F-score @ 1\%}} \\
\midrule
3D-R2N2 \cite{choy20163d}        
    & 0.351     & 0.368     & 0.372     & 0.378    
    & 0.382         & 0.383     & 0.382         & 0.382   &   0.383
    \\
AttSets \cite{yang2020robust}       
    & 0.395     & 0.418     & 0.426     & 0.430    
    & 0.432         & 0.444     & 0.445         & 0.447   &   0.448
    \\
Pix2Vox/F \cite{xie2019pix2vox}   
    & 0.364     & 0.393     & 0.404     & 0.409    
    & 0.412         & 0.417     & 0.420         & 0.423   &   0.423
    \\
Pix2Vox/A \cite{xie2019pix2vox}     
    & 0.405     & 0.435     & 0.448     & 0.449    
    & 0.452         & 0.456     & 0.458         & 0.459   &   0.460
    \\
Pix2Vox++/F \cite{xie2020pix2vox++} 
    & 0.394     & 0.422     & 0.432     & 0.437    
    & 0.440         & 0.446     & 0.449         & 0.450   &   0.451
    \\
Pix2Vox++/A \cite{xie2020pix2vox++} 
    & \textbf{0.436}  & \textbf{0.452}   & \textbf{0.455}   & \textbf{0.457}
    & \textbf{0.458}      & 0.459  & 0.460  & 0.461 & 0.462
    \\
LegoFormer-M 
    & 0.282  & 0.392        & 0.428     & 0.444   
    & 0.453      & \textbf{0.464}        & \textbf{0.470}         & \textbf{0.472}   &   \textbf{0.473}
    \\
LegoFormer-S 
    & 0.404  & -   & -   & -    
    & -      & -     & -    &   -    &   -
    \\
    \bottomrule
\end{tabular}}
\caption{Comparison of multi-view reconstruction between models on ShapeNet at $32^3$ resolution. Mean IoU and F-Score@1\% reported for all categories.}
\label{tab:multi_view_comparison}
\end{table*}

\section{Experiments}

\subsection{Evaluation protocol and implementation details}

\textbf{Dataset:}
Following \cite{choy20163d, xie2019pix2vox, xie2020pix2vox++} we use rendered images from ShapeNet \cite{wu20153d} to evaluate the proposed method. To have comparable results, we follow the settings of \cite{choy20163d} and use a subset of ShapeNet, which includes 43783 models from 13 categories. Each model is rendered from 24 different poses. The original images of size $137 \times 137$ are resized to $224 \times 224$ and a uniform background color is applied before passing them to the network. The ground truth targets are 3D occupancy grids with size $32^3$ and target volumes aligned to a canonical reference frame. 

\textbf{Metrics:}
3D Intersection over Union (IoU) and F-score are used to measure the reconstruction performance. Given predicted $\mathcal{R}$ and ground-truth $\mathcal{G}$ occupancy volumes, the former measures the ratio of intersecting voxels from both volumes to their union. The ratio ensures that the calculation is object size independent. The latter metric, proposed by \cite{knapitsch2017tanks}, focuses on the quality of the surface reconstruction and measures the percentage of points from the object surfaces that are closer than a predefined threshold. We follow the setup of \cite{xie2020pix2vox++} to convert the predicted volumes to point clouds and use the implementation provided by \cite{tatarchenko2019single} for calculating the F-score. For both metrics, a higher value means better reconstruction.

\textbf{Implementation Details:}
Both LegoFormer-M and LegoFormer-S are trained with batch size 128 using $224 \times 224$ RGB images as input and $32^3$ voxelized reconstructions as output. The models are implemented using PyTorch-Lightning \cite{falcon2019pytorch} and trained on Nvidia A100 and T4 GPUs on Google Cloud Platform (GCP) using Adagrad optimizer \cite{duchi2011adaptive}. The learning rate is set to 0.01 with 10K warmup steps. For multi-view models, the number of input views is fixed to 8 during the training as we experimentally found that fixing the view count helps to achieve higher performance than updating it between steps (experiments reported in the Supplementary \cref{section:view_count_ablation}). The views are randomly sampled out of 24 views at each iteration.
The number of layers for both encoder and decoder is set to 8. The transformer dimensionality is set to 768 and 4096 for input token and feed-forward layer, respectively. After running ablation studies with different number of decoder queries, the decoder input count is fixed to 12, meaning that the reconstructed volume is formed from 12 rank-1 estimations. At inference time after the sum-aggregation and clipping, we use a threshold $\tau=0.3$ to obtain the occupancy grid. LegoFormer-M is trained for 80K steps, while LegoFormer-S is trained for 160K steps. No learning rate decay was used. Obtaining the reconstructed volumes from the decomposition factors is implemented using Einstein Summation for efficient use of the memory. 
We use data and implementations along with pre-trained models provided by authors under the MIT License for comparison to other works. 
Our code and trained models will be made publicly available.

\subsection{Multi-view Reconstruction}

We start by evaluating the multi-view reconstruction performance of LegoFormer-M against 3D-R2N2 \cite{choy20163d}, AttSets \cite{yang2020robust}, Pix2Vox \cite{xie2019pix2vox}, and Pix2Vox++ \cite{xie2020pix2vox++}. As shown in \cref{tab:multi_view_comparison} the proposed method slightly outperforms other techniques when more than 12 views are given. The difference in F-score is higher than the difference in IoU, meaning that LegoFormer reconstructs object surfaces better. The performance for a lower number of views is comparable to the state-of-the-art method Pix2Vox++ and the method outperforms 3D-R2N and AttSets when two or more views are given. Increasing the number of input views increases the performance of all methods, however for our method the gain is higher thanks to the use of the attention mechanism during the view encoding. For example, we get a $+12.5$ mIoU between not using encoder side attention (1 view) and using it (2 views), showing the impact of this component. 

\begin{figure}[t]
    \centering 
    \includegraphics[page=1]{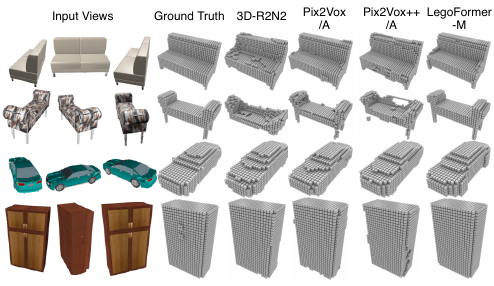}
    \caption{Multi-view object reconstructions for 4 input views (only 3 are shown) on ShapeNet with $32^3$ resolution.
    \vspace{-0.5cm}}
    \label{fig:multi_view_comparison}
\end{figure}

\cref{fig:multi_view_comparison} displays examples of reconstructions obtained from 4 input views (only 3 are shown). In general, the reconstructions by LegoFormer tend to have less noise and smoother surfaces. This property can be attributed to the constraints that our tensor decomposition imposes. Since the output is an aggregation of rank-1 tensors obtained as the cross-product between three vectors, it is harder for the network to predict random spike-like voxels and non regular surfaces. This formulation naturally acts as a regularizer that forces to reconstruct the object from well-connected chunks. Qualitative examples of this are the sitting and the backrest areas of the bench and sofa reconstructions.

\subsection{Comparison to Multi-view 3D Reconstruction with Transformer \cite{wang2021multi}}

Concurrently with our work, Wang et al. \cite{wang2021multi} developed a similar model based on a transformer architecture for multi-view reconstruction. In \cite{wang2021multi}, the encoder side of the transformer takes full views as input, as we do, but uses a different attention mechanism to foster different representations for each view. On the decoder side, instead, they stick to a more vanilla formulation and use a decoding schema similar to our Naive-nAR variant described in \cref{sec:alternative_schemes}. 
Unfortunately, the authors of \cite{wang2021multi} did not release their code nor their trained models, making an extensive comparison challenging. We tried to replicate their results following the description on the paper, but unfortunately we were not able to achieve the good performance reported. 
As such we present here only a preliminary comparison between the two works, with the caveat of taking the performance for \cite{wang2021multi} directly from the paper. 
In \cref{tab:concurrent_work_comparison} we compare three variants of the method proposed in \cite{wang2021multi} (rows 1,2,3) and our model. Our formulation is competitive or better than \cite{wang2021multi} for a low view count, while for an high view count we perform slightly worse than EVoIT but better than the other variants. Among the three variants of \cite{wang2021multi} EVoIT is the only one using an enhanced attention schema in the encoder part of the network, one of the main contributions of \cite{wang2021multi}.
We believe that these experimental results show how our decoding schema is superior to the one used in \cite{wang2021multi} while their enhanced attention helps in the case of many views provided as input. 
%
Combining the strength of both methods is an exciting future development for this branch of works, however, as mentioned, replicating the result of EVoIT has proven challenging.
Finally, while \cite{wang2021multi} proposes models optimized for inference mainly on many input views, we preferred to develop methods with more consistent performance also on a low number of views, getting to the extreme of LegoFormer-S which is explicitly designed to achieve good performance with a single input view and which we are going to evaluate next.


  


\setlength{\tabcolsep}{3pt} 
\renewcommand{\arraystretch}{1} 

\begin{table}[]
  \centering
\scalebox{0.85}{
\begin{tabular}{l|ccccccccc}
            \toprule
            & \multicolumn{7}{c}{Evaluation view count} \\
            \cmidrule(r){2-8}
            Model & 4 & 6 & 8 & 12 & 16 & 20 & 24 \\
            \midrule
            \textbf{IoU} \\
            \midrule
VolT \cite{wang2021multi}    
    & 0.605 & 0.662 & 0.681 & 0.699 & 0.706 & 0.711 & 0.714 \\
VolT+ \cite{wang2021multi}
    & 0.695 & 0.704 & 0.707 & 0.711 & 0.714 & 0.715 & 0.716 \\
EVolT \cite{wang2021multi}
    & 0.609 & 0.675 & 0.698 & \textbf{0.720} & \textbf{0.729} & \textbf{0.735} & \textbf{0.738} \\
LegoFormer-M
    & \textbf{0.694} & \textbf{0.709} & \textbf{0.713} & 0.717 & 0.719 & 0.721 & 0.721 \\
            \midrule
            \textbf{F-Score@1\%} \\
            \midrule
VolT \cite{wang2021multi}    
    & 0.356 & 0.410 & 0.430 & 0.450 & 0.459 & 0.464 & 0.468 \\
VolT+ \cite{wang2021multi}
    & \textbf{0.451} & \textbf{0.460} & \textbf{0.464} & 0.469 & 0.472 & 0.474 & 0.475 \\
EVolT \cite{wang2021multi}
    & 0.358 & 0.423 & 0.448 & \textbf{0.475} & \textbf{0.486} & \textbf{0.492} & \textbf{0.497} \\
LegoFormer-M
    & 0.444 & \textbf{0.460} & \textbf{0.464} & 0.470 & 0.472 & 0.473 & 0.474 \\
    \bottomrule
\end{tabular}
}

\caption{Comparison to the concurrent work by Wang et al. \cite{wang2021multi} including three variations of their proposed architecture.}
\label{tab:concurrent_work_comparison}
\end{table}

\subsection{Single-view Reconstruction}

\begin{figure}
    \centering
    \includegraphics[page=1]{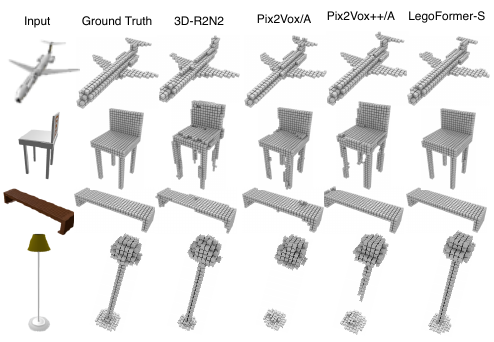}
    \caption{Single-view object reconstructions on ShapeNet with $32^3$ resolution.}
    \label{fig:single_view_comparison}
\end{figure}

The first column of \cref{tab:multi_view_comparison} compares LegoFormer-S and LegoFormer-M against state-of-the-art methods for single-view object reconstruction. As expected LegoFormer-S, which uses image patches as input tokens, outperforms LegoFormer-M, which uses the full image as a single token, showing again the advantage of using the attention mechanism on the encoder. LegoFormer-S shows superior performance with respect to all competitors except for Pix2Vox++/A. Some examples of single view reconstruction are shown in \cref{fig:single_view_comparison}, highlighting how, once again, the shapes predicted by LegoFormer-S are much smoother and more regular than the one predicted by the competitors. A category-wise comparison to the other single-view reconstruction techniques is reported in the Supplementary \cref{appendix:single_view_performance}.

\subsection{Single-view reconstruction from real images}

\begin{figure}
    \centering
    \includegraphics[width=\columnwidth]{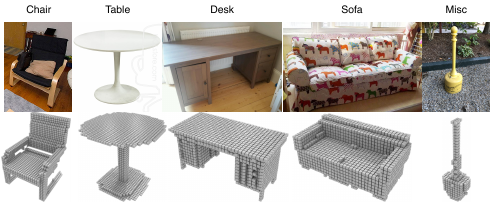}
    \caption{Reconstruction of real-world objects from Pix3D \cite{sun2018pix3d} obtained by a LegoFormer-S model trained only on ShapeNet \cite{chang2015shapenet}. The four leftmost samples represent categories present in ShapeNet, while the \emph{Misc.} one is unknown to the model.}
    \label{fig:pix3d}
\end{figure}

To test the generalization performance of our method, we test it on real-world settings using the Pix3D dataset \cite{sun2018pix3d}. 
Pix3D provides a single view for various real objects together with a segmentation masks. 
For each object we use the mask to segment out the background and replace it with a constant color. 
The resulting images are provided as input to a LegoFormer-S trained only on Shapenet and we report examples of the predicted models for different categories in \cref{fig:pix3d}.
In particular the four leftmost samples represent categories overlapping the ShapeNet ones, while the \emph{Misc.} sample represent a completely unseen category. 
The quality of the reconstructions highlights how our method can successfully generalize across domains (synthetic-to-real) and, partially, also to completely unseen categories (\emph{Misc.}).
More results are reported in the supplementary material.

\subsection{Model Analysis}

\begin{figure*}
  \centering
  \includegraphics[page=1]{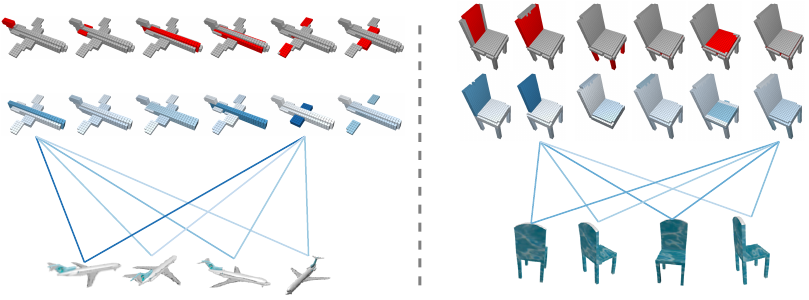}
  \caption{Visualization of the decoder-to-decoder (top) and decoder-to-encoder (bottom) attention for a LegoFormer-M model trained with 6 queries. Red color shows the part being reconstructed. Shades of blue  denotes the attention score with darker meaning higher. The decoder-to-encoder attention is visualized for two queries for simplicity.}
  \label{fig:multiview_attention}
\end{figure*}

During the development of this work we found out that we can get insight on the reconstruction process of LegoFormer by analyzing and visualizing the attention scores. Our architecture uses 3 types of attention: encoder-to-encoder, decoder-to-encoder and decoder-to-decoder. The second gives information on where the network "looks" when reconstructing specific parts. The last identify which parts are taken into account when decoding a specific query.

\cref{fig:multiview_attention} visualizes the attention scores for two examples from a LegoFormer model trained with 6 queries. We use a reduced number of queries for this experiment to ease the visualization. First, the network pays more attention to the airplane body while reconstructing its tail, likely because the tail can be predicted by looking at the plane's body. Second, when reconstructing a specific part, the surrounding parts get more attention, e.g.,  the last three columns of the airplane and the first two for the chair.
Also, parts split in multiple pieces, like the airplane wings and chair backrest have significantly higher attention for the other adjacent pieces.
Consistency could be an explanation for these observations, i.e., the network tries to make sure that predicted parts fit well together.
Last but not least, one of the input views usually gets the majority of the attention from the decoder side. 
This suggests that the model is focusing on a single output of the encoder as "reference" to guide the reconstruction, while the others are mainly used to refine details. Indeed, thanks to the attention layers, each encoder outputs represent the corresponding input view, but also aggregates information coming from the others. 


\begin{figure}
  \centering
  \includegraphics[width=\columnwidth]{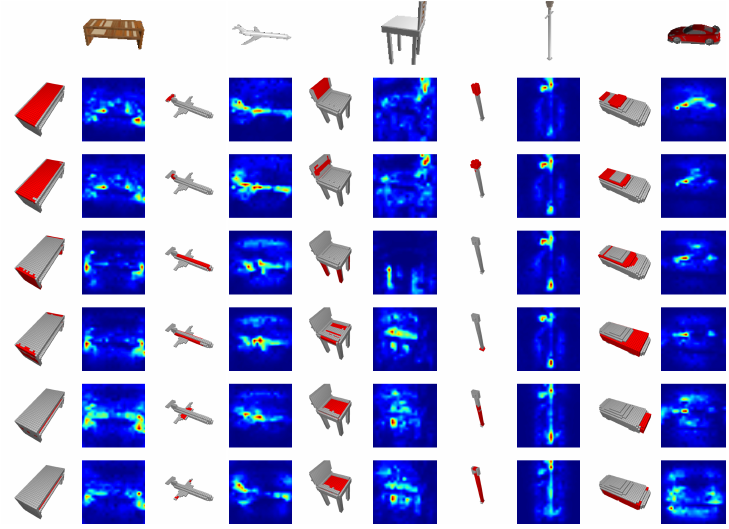}
  \caption{Decoder-to-Encoder attention for LegoFormer-S trained with input patch size $1^2$. The input image is shown in the top row, odd columns show the part being reconstructed by each query (in red), while even columns plot the attention on the input tokens with a jet color map. 
  \vspace{-0.5cm}
  }
  \label{fig:single_view_attention}
\end{figure}


We also show the encoder-decoder attention for LegoFormer-s in \cref{fig:single_view_attention}. In LegoFormer-S the decoder attending to individual outputs of the encoder focuses on particular patches of the input view, therefore we opted for a heat-map to plot the attention.
To enhance the visualization we trained a special version of LegoFormer-S using input patches of size $1^2$. The attention maps give insights into where the network "looks" when predicting a particular block.
First, as expected, the attention is always higher on the object and surrounding parts, likely to understand the silhouette of the object. 
Second, the attention can be focused on a particular place or distributed over a larger area. 
For example, while predicting the tail of the plane (first and second row), the attention is spread all over the plane's body. In contrast, while reconstructing the chair legs (third row), it is very focused on that part only. 
The reason behind this difference could be that the shape of the plane tails have higher variance, and "looking" to other parts is helpful, while the chair legs are more or less similar to each other and self-sufficient for the reconstruction. Lastly, the decoder attends to similar parts of the images when reconstructing nearby blocks. For example, when reconstructing the central parts of the sitting of the chair (last two rows) the attention maps generated look quite similar, showing a strong relationship between the shape being generated and on which part of the input view the model decides to attend.

\begin{figure}[t]
  \centering
  \includegraphics[page=1]{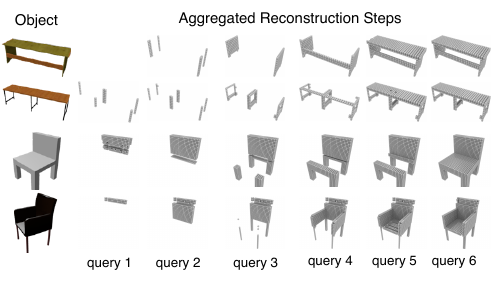}
    \caption{Reconstruction from a model trained with 6 queries on Shapenet. We show aggregated results for each query, i.e., the model above query 2 represent the output of query 1 + query 2.
    }
    \label{fig:progressive_aggregation}
\end{figure}

Finally, during our experiments we always observed intra- and inter- category consistency in how the queries get decoded into object parts. \cref{fig:progressive_aggregation} displays incremental reconstructions obtained aggregating decoded queries for 2 categories each with 2 examples. For both tables, the first queries predict the bottom parts, while the last ones predict the top. A similar pattern exists for chairs - the backrest gets predicted by the first queries followed by the legs and the seat. 
Furthermore, we observe some form of consistency between categories: the first queries tend to predict the bottom and sides of the object while the lasts predict the top part and the inside. 
These experimental observations suggest that during the training, each query specialize on specific parts that are similar between objects from the same and different categories.


\subsection{Comparison to other decoding schemes}
\label{sec:other_schemas_experimental}
\setlength{\tabcolsep}{4pt} 
\renewcommand{\arraystretch}{1} 
\begin{table}[t]
  \centering
 \scalebox{0.9}{
\begin{tabular}{l|cccc|cccc}
        \toprule
        & \multicolumn{4}{c|}{IoU} &  \multicolumn{4}{c}{F-Score@1\%} \\
        \cmidrule(r){2-5} \cmidrule(r){6-9}
        \rot{Views} 
         & \rot{Naive} 
         & \rot{Naive-nAR} 
         & \rot{Naive-Full} 
         & \rot{LegoFormer-M} 
         & \rot{Naive} 
         & \rot{Naive-nAR} 
         & \rot{Naive-Full} 
         & \rot{LegoFormer-M} \\
         \midrule
1   & 0.500 & 0.557     & 0.558      & \textbf{0.617}        & 0.290 & 0.333     & 0.337      & \textbf{0.364}        \\
2  & 0.552 & 0.640     & 0.643      & \textbf{0.674}        & 0.339 & \textbf{0.427}     & 0.426      & 0.422        \\
3  & 0.567 & 0.663     & 0.659      & \textbf{0.689}        & 0.353 & \textbf{0.456}     & 0.447      & 0.438        \\
4  & 0.573 & 0.670     & 0.668      & \textbf{0.695}        & 0.359 & \textbf{0.464}     & 0.458      & 0.445        \\
5  & 0.577 & 0.676     & 0.672      & \textbf{0.699  }      & 0.363 & \textbf{0.472}     & 0.463      & 0.449        \\
8  & 0.582 & 0.681     & 0.675      & \textbf{0.704}        & 0.367 & \textbf{0.480}     & 0.467      & 0.455        \\
12 & 0.583 & 0.685     & 0.680      & \textbf{0.706}        & 0.369 & \textbf{0.486}     & 0.474      & 0.457        \\
16 & 0.584 & 0.686     & 0.681      & \textbf{0.707}        & 0.371 & \textbf{0.489}     & 0.477      & 0.459        \\
20 & 0.585 & 0.687     & 0.681      & \textbf{0.708}        & 0.372 & \textbf{0.489}     & 0.476      & 0.459        \\
\bottomrule

\end{tabular}
}
\caption{Comparison of multi-view reconstruction performance of the decoding schemes on ShapeNet at $32^3$ resolution. Mean IoU and F-Score@1\% reported for all categories. All models are trained with 4 input views.
\vspace{-0.5cm}
}
\label{tab:comparison_decoding_schemes}
\end{table}

To experimentally validate our design choices we compare the four decoding schemes described in \cref{sec:alternative_schemes} in \cref{tab:comparison_decoding_schemes}. 
All models are trained on 4 input views. LegoFormer achieves the highest IoU in all view counts while both "Naive-nAR" and "Naive-Full" have higher F-score performance suggesting that the former is better for objects with solid interior while the latter are good at reconstructing surfaces. The "Naive-nAR" decoding scheme constantly achieves the highest F-score except for a single view case. Compared to LegoFormer, it solves an easier task where the subspace predicted by each query is predefined and doesn't change. However, the memory requirement is much higher, making it a less attractive solution when considering scaling the output resolution. Finally, "Naive-Full" is the most canonical architecture directly predicting each voxel independently. This solution is still competitive but suffers from the same scalability problem of "Naive-nAR" when increasing the output resolution. 
The "Naive" model has the worst performance, which can be attributed to accumulated error. As the predictions are made in sequence while conditioning on previous steps, a slight mistake at some early steps will result in wrong predictions later on. 
We observed a huge gap between train and test performance for this model, leading us to conclude that autoregressive decoding schemes are not good candidates for 3D voxel grid reconstruction. 
An additional qualitative comparison of the decoding schemes is provided in the Supplementary \cref{appendix:decoding_schemes_qualitative_comparison}.

\section{Conclusion \& Discussion}

We presented LegoFormer, a transformer model to reconstruct an object from multiple views via its decomposition factors. 
In contrast to previous works, the proposed method combines view encoding and volume prediction under the same network and allows information sharing among views at every stage.
We also presented extensive insights on transformer-based decoding schemes for 3D reconstruction and showed the advantages of the reconstruction via attention-based tensor decomposition on synthetic and real dataset.

Finally, several limitations of this work should be considered in the future. The output scheme has been explored only for the occupancy grids. While in theory, it can be extended to signed distance fields with minimal changes. Furthermore, the performance was measured regarding the output resolution $32^3$. The method can be adapted to higher resolutions by only changing the output dimensionality or by integrating implicit refinement network like \cite{chibane20ifnet}.


\section{Acknowledgements}
We thank Diego Martin Arroyo and Janis Postels for the discussions. We are also grateful to Google University Relationship GCP Credit Program for the support of this work by providing computational resources.

{\small
\bibliographystyle{ieee_fullname}
\bibliography{main}
}

\newpage

\begin{appendices}

In this supplementary material we report additional experiments performed on ShapeNet\cite{chang2015shapenet}: ablation studies on hyper parameters of our model in \cref{appendix:ablation}, an extended evaluation of LegoFormer for single view reconstruction broken down by category in \cref{appendix:single_view_performance} and more results comparing the different explored decoding schema in \cref{appendix:decoding_schemes_qualitative_comparison}.
We also show more generalization results on the challenging Pix3D\cite{sun2018pix3d} dataset in \cref{appendix:incremental_real} and extensively discuss limitations of the method in \cref{appendix:limitations}.

\section{Additional ablation tests}
\label{appendix:ablation}
We report here additional ablation studies to highlight the effect of different hyperparameters on the performance of the proposed architectures.
    
\subsection{Number of Decomposition Factors}
\label{section:query_count_ablation}
In LegoFormer, the output volume is formed as a sum of $n$ rank-1 approximations obtained by independently processing $n$ learned queries. The number of approximations considered is correlated to the reconstruction performance achievable. \cref{tab:view_count_vs_query_count} lists the performance of models trained with a different number of output queries against a different number of input views considered at test time. For each query count, the model was trained from scratch. As it can be easily observed, increasing the number of queries corresponds to an increase in performance—however, however the increase stalls after 12 queries. Interestingly, with only 2 queries a good IoU and F1-score performance can be achieved, only around 5 points lower than the best configurations. This means that our architecture can approximate very complex object shapes using as few as two rank-1 approximations.

\setlength{\tabcolsep}{6pt} 
\renewcommand{\arraystretch}{1.1} 

\begin{table}[htbp]
    \centering
    
\scalebox{0.82}{
\begin{tabular}{lccccccc}
    \toprule
    & \multicolumn{7}{c}{Query count} \\
        \cmidrule(r){2-8}
    Views & 2 & 3 & 4 & 6 & 8 & 12 & 16 \\
    
\midrule
\multicolumn{8}{l}{\textbf{Metric: IoU} } \\
\midrule

\multicolumn{1}{l}{1} \vline
    & 0.585     & 0.598     & 0.606     & 0.609     & 0.611     & \textbf{0.617}      & 0.614      \\
\multicolumn{1}{l}{2} \vline
    & 0.627     & 0.648     & 0.656     & 0.663     & 0.668     & \textbf{0.674}      & \textbf{0.674}      \\
\multicolumn{1}{l}{3} \vline
    & 0.637     & 0.661     & 0.669     & 0.679     & 0.684     & \textbf{0.689}      & \textbf{0.689}      \\
\multicolumn{1}{l}{4} \vline
    & 0.643     & 0.667     & 0.675     & 0.685     & 0.690     & \textbf{0.695}      & \textbf{0.695}      \\
\multicolumn{1}{l}{5} \vline
    & 0.646     & 0.670     & 0.679     & 0.689     & 0.694     & \textbf{0.699}      & \textbf{0.699}      \\
\multicolumn{1}{l}{8} \vline
    & 0.650     & 0.675     & 0.684     & 0.694     & 0.700     & \textbf{0.704}      & \textbf{0.704}      \\
\multicolumn{1}{l}{12} \vline
    & 0.652     & 0.677     & 0.685     & 0.697     & 0.702     & 0.706      & \textbf{0.707}      \\
\multicolumn{1}{l}{16} \vline
    & 0.653     & 0.678     & 0.687     & 0.698     & 0.703     & 0.707      & \textbf{0.708}      \\
\multicolumn{1}{l}{20} \vline
    & 0.653     & 0.679     & 0.687     & 0.699     & 0.704     & 0.708      & \textbf{0.709}      \\
    
\midrule
\multicolumn{8}{l}{\textbf{Metric: F-score @ 1\%} } \\
\midrule

\multicolumn{1}{l}{1} \vline
    & 0.323     & 0.336     & 0.345     & 0.355     & 0.358     & \textbf{0.364}      & 0.362      \\
\multicolumn{1}{l}{2} \vline
    & 0.362     & 0.383     & 0.392     & 0.407     & 0.413     & \textbf{0.422}      & 0.421      \\
\multicolumn{1}{l}{3} \vline
    & 0.373     & 0.397     & 0.407     & 0.424     & 0.429     & \textbf{0.438}      & \textbf{0.438}      \\
\multicolumn{1}{l}{4} \vline
    & 0.379     & 0.402     & 0.413     & 0.430     & 0.436     & \textbf{0.445}      & 0.444      \\
\multicolumn{1}{l}{5} \vline
    & 0.382     & 0.406     & 0.417     & 0.434     & 0.441     & \textbf{0.449}      & \textbf{0.449}      \\
\multicolumn{1}{l}{8} \vline
    & 0.386     & 0.411     & 0.422     & 0.440     & 0.447     & \textbf{0.455}      & \textbf{0.455}      \\
\multicolumn{1}{l}{12} \vline
    & 0.389     & 0.414     & 0.424     & 0.443     & 0.451     & 0.457      & \textbf{0.458}      \\
\multicolumn{1}{l}{16} \vline
    & 0.389     & 0.415     & 0.425     & 0.445     & 0.452     & \textbf{0.459}      & \textbf{0.459}      \\
\multicolumn{1}{l}{20} \vline
    & 0.390     & 0.415     & 0.426     & 0.445     & 0.453     & 0.459      & \textbf{0.461}     \\
\bottomrule
\end{tabular}
}
\caption{Comparison of the reconstruction performance with respect to the number of queries used on the decoder side. For each number of queries a separate model was trained from scratch.}
\label{tab:view_count_vs_query_count}
\end{table}

\subsection{Training with different view counts}\label{section:view_count_ablation}

We also study the effect of the number of input views considered during training the LegoFormer-M model. We considered both using a fixed view count (2, 4 or 8) as well as random sampling the number of input views considered at every training iteration. \cref{tab:training_view_counts} reports the results of these experiments. 
First of all, we can notice how, for the fixed number of inputs, there is a loose correlation between the number of training views and the performance at test time. Indeed, a model trained with few inputs performs better when running inference on few views and vice versa. This can be observed comparing the performance of the model trained with 4 and 8 views, the former outperforms the latter for any inference views count lower than 4, but after this threshold, the latter model performs better.
By training with a dynamic number of views, we tried to achieve higher performance regardless of view count during the evaluation. Unfortunately, we observed harder convergence with this setup resulting in lower performance. This could be linked to the instability problems of the transformer training \cite{liu2020understanding}. We plan to investigate further this aspect of the model in future works.
    
\setlength{\tabcolsep}{6pt} 
\renewcommand{\arraystretch}{1} 

\begin{table*}[]
  \centering
  
\scalebox{0.9}{
\begin{tabular}{c|ccccccccc}
        \toprule
        & \multicolumn{9}{c}{Evaluation view count} \\
        \cmidrule(r){2-10}
        \makecell{Training\\view count}
        & 1 & 2 & 3 & 4 & 5 & 8 & 12 & 16 & 20 \\
\midrule
\textbf{Metric: IoU} \\
\midrule
$2$   &
    0.635 & 0.668 & 0.677 & 0.681 & 0.684 & 0.688 & 0.690 & 0.691 & 0.691 \\
$4$   &
    0.617 & 0.674 & 0.689 & 0.695 & 0.699 & 0.704 & 0.706 & 0.707 & 0.708 \\
$8$   &
    0.519 & 0.644 & 0.679 & 0.694 & 0.703 & 0.713 & \textbf{0.717} & \textbf{0.719} & \textbf{0.721} \\
$[1, 10)$   &
    0.601 & 0.638 & 0.652 & 0.659 & 0.663 & 0.669 & 0.672 & 0.673 & 0.674 \\
\cdashlinelr{1-10}
Pix2Vox++/A \cite{xie2020pix2vox++} &
    \textbf{0.670} & \textbf{0.695} & \textbf{0.704} & \textbf{0.708} & \textbf{0.711} & \textbf{0.715} & \textbf{0.717} & 0.718 & 0.719 \\
\midrule
\textbf{Metric: F-score @ 1\%} \\
\midrule
$2$   &
    0.377 & 0.412 & 0.421 & 0.426 & 0.428 & 0.433 & 0.435 & 0.436 & 0.437 \\
$4$   &
    0.364 & 0.422 & 0.438 & 0.445 & 0.449 & 0.455 & 0.457 & 0.459 & 0.459 \\
$8$   &
    0.282 & 0.392 & 0.428 & 0.444 & 0.453 & \textbf{0.464} & \textbf{0.470} & \textbf{0.472} & \textbf{0.473} \\
$[1, 10)$   &
    0.349 & 0.384 & 0.398 & 0.404 & 0.409 & 0.415 & 0.418 & 0.419 & 0.420 \\
\cdashlinelr{1-10}
Pix2Vox++/A \cite{xie2020pix2vox++} &
    \textbf{0.436} & \textbf{0.452} & \textbf{0.455} & \textbf{0.457} & \textbf{0.458} & 0.459 & 0.460 & 0.461 & 0.462 \\
\bottomrule
\end{tabular}
}

\caption{Effect of the number of input views used during training on the inference-time performance on the ShapeNet dataset at $32^3$ resolution. In the setup of the column $[1, 10)$ view count is changed at every iteration to a number between 1 and 9, inclusive. We also report the performance of the strongest competitor, Pix2Vox++/A \cite{xie2020pix2vox++} as a reference.}
  \label{tab:training_view_counts}
\end{table*}

\subsection{Reducing model dimensionality}

The parameter count of the transformer models can grow pretty quickly, resulting in higher memory requirements. A number of techniques were proposed to counteract the parameter growth of transformer architectures. One of the techniques is to share weights between transformer encoder and decoder layers, as proposed in ALBERT \cite{lan2019albert}. Using this technique, only a single transformer layer is defined for the encoder and decoder, and the layer is repeatedly applied to the input.
We experiment with this technique and were able to reduce the model size by 5.5 times, from 168M to 30.6M parameters, while experiencing minor performance drop as reported in \cref{tab:weight_sharing}. Based on this experiment, we show that there exist a nice trade-off between memory requirement and performance for LegoFormer architectures, and the former can be decreased significantly without affecting the latter much.

\setlength{\tabcolsep}{6pt} 
\renewcommand{\arraystretch}{1} 

\begin{table*}[htbp]
  \centering
  
\scalebox{1.0}{
\begin{tabular}{l|ccc|ccc}
            \toprule
            & \multicolumn{3}{c|}{IoU} &  \multicolumn{3}{c}{F-Score@1\%} \\
            \cmidrule(r){2-4} \cmidrule(r){5-7}
            \diagbox{View\\Count}{\makecell{Setup}} 
            & Vanilla & \makecell{Parameter\\Sharing} & Difference
            & Vanilla & \makecell{Parameter\\Sharing} & Difference
            \\
            \midrule
1 view  
    & \textbf{0.617}	    & 0.596     & -0.021
    & \textbf{0.364}     & 0.345     & -0.019
    \\ 
2 views 
    & \textbf{0.674}	    & 0.661     & -0.013
    & \textbf{0.422}     & 0.406     & -0.016
    \\
3 views 
    & \textbf{0.689}	    & 0.678     & -0.011
    & \textbf{0.438}     & 0.424     & -0.014
    \\
4 views 
    & \textbf{0.695}	    & 0.685     & -0.010
    & \textbf{0.445}     & 0.432     & -0.013
    \\
5 views 
    & \textbf{0.699}	    & 0.691     & -0.008
    & \textbf{0.449}    & 0.438     & -0.011
    \\
8 views 
    & \textbf{0.704}	    & 0.697     & -0.007
    & \textbf{0.455}     & 0.444     & -0.011
    \\
12 views 
    & \textbf{0.706}	    & 0.700     & -0.006
    & \textbf{0.457}     & 0.448     & -0.009
    \\
16 views 
    & \textbf{0.707}	    & 0.701     & -0.006
    & \textbf{0.459}     & 0.449     & -0.010
    \\
20 views 
    & \textbf{0.708}	    & 0.702     & -0.006
    & \textbf{0.459}     & 0.450     & -0.009
    \\
    \bottomrule
\end{tabular}
}

\caption{Comparison between a vanilla LegoFormer and a parameter sharing LegoFormer on the ShapeNet dataset. In the latter case, all layers of the encoder and decoder use the same parameters, meaning that a single layer is repeatedly applied for each side resulting in using $5.5x$ less parameters.}
  \label{tab:weight_sharing}
\end{table*}

\section{Single view performance per category}
\label{appendix:single_view_performance}

We extend the single view evaluation of LegoFormer on ShapeNet\cite{chang2015shapenet} in \cref{tab:comparison_single_view} where we compare it against 10 recent state-of-the-art methods that predict voxel grids \cite{xie2019pix2vox, xie2020pix2vox++, choy20163d, tatarchenko2017octree, richter2018matryoshka}, triangle meshes \cite{groueix2018papier, wang2018pixel2mesh} or implicit representations \cite{mescheder2019occupancy, chen2019learning}. 
We report both category-wise performance and "overall", i.e., the average over the available classes for each method. In almost every category, LegoFormer-S shows competitive performance against the other methods. Also, LegoFormer-S performs significantly better than LegoFormer-M in all categories, showing the importance of using attention-based encoding. Overall, the proposed method is competitive with recent proposals in the literature, outperforming several of them.
    
\setlength{\tabcolsep}{6pt} 
\renewcommand{\arraystretch}{1} 
\begin{sidewaystable*}[htbp]
  \centering
  \small
\begin{tabular}{l|ccccccccccccccc}
\diagbox{Category}{Method}
& \rot{3D-R2N2 \cite{choy20163d}} 
& \rot{OGN \cite{tatarchenko2017octree}}   
& \rot{Matryoshka \cite{richter2018matryoshka}} 
& \rot{AtlasNet \cite{groueix2018papier}} 
& \rot{Pixel2Mesh \cite{wang2018pixel2mesh}} 
& \rot{OccNet \cite{mescheder2019occupancy}} 
& \rot{IM-Net \cite{chen2019learning}} 
& \rot{AttSets \cite{yang2020robust}} 
& \rot{Pix2Vox/F \cite{xie2019pix2vox}} 
& \rot{Pix2Vox/A \cite{xie2019pix2vox}} 
& \rot{Pix2Vox++/F \cite{xie2020pix2vox++}} 
& \rot{Pix2Vox++/A \cite{xie2020pix2vox++}}
& \rot{LegoFormer-M} 
& \rot{LegoFormer-S} \\
    \midrule
    \textbf{IoU} \\
    \midrule
airplane   & 0.513   & 0.587 & 0.647      & 0.493    & 0.508      & 0.532  & \textbf{0.702}  & 0.594   & 0.600     & 0.684     & 0.607       & 0.674       & 0.480        & 0.641        \\
bench      & 0.421   & 0.481 & 0.577      & 0.431    & 0.379      & 0.597  & 0.564  & 0.552   & 0.538     & \textbf{0.616}     & 0.544       & 0.608       & 0.326        & 0.612        \\
cabinet    & 0.716   & 0.729 & 0.776      & 0.257    & 0.732      & 0.674  & 0.680  & 0.783   & 0.765     & 0.792     & 0.782       & \textbf{0.799}       & 0.640        & 0.780        \\
car        & 0.798   & 0.828 & 0.850      & 0.282    & 0.670      & 0.671  & 0.756  & 0.844   & 0.837     & 0.854     & 0.841       & \textbf{0.858}       & 0.724        & 0.852        \\
chair      & 0.466   & 0.483 & 0.547      & 0.328    & 0.484      & 0.583  & \textbf{0.644}  & 0.559   & 0.535     & 0.567     & 0.548       & 0.581       & 0.442        & 0.557        \\
display    & 0.468   & 0.502 & 0.532      & 0.457    & 0.582      & \textbf{0.651}  & 0.585  & 0.565   & 0.511     & 0.537     & 0.529       & 0.548       & 0.391        & 0.524        \\
lamp       & 0.381   & 0.398 & 0.408      & 0.261    & 0.399      & \textbf{0.474}  & 0.433  & 0.445   & 0.435     & 0.443     & 0.448       & 0.457       & 0.428        & 0.453        \\
speaker    & 0.662   & 0.637 & 0.701      & 0.296    & 0.672      & 0.655  & 0.683  & 0.721   & 0.707     & 0.714     & 0.721       & \textbf{0.721}       & 0.645        & 0.712        \\
rifle       & 0.544   & 0.593 & 0.616      & 0.573    & 0.468      & 0.656  & \textbf{0.723}  & 0.601   & 0.598     & 0.615     & 0.594       & 0.617       & 0.465        & 0.632        \\
sofa       & 0.628   & 0.646 & 0.681      & 0.354    & 0.622      & 0.669  & 0.694  & 0.703   & 0.687     & 0.709     & 0.696       & \textbf{0.725}       & 0.483        & 0.707        \\
table      & 0.513   & 0.536 & 0.573      & 0.301    & 0.536      & \textbf{0.659}  & 0.621  & 0.590   & 0.587     & 0.601     & 0.609       & 0.620       & 0.489        & 0.596        \\
telephone  & 0.661   & 0.702 & 0.756      & 0.543    & 0.762      & 0.794  & 0.762  & 0.743   & 0.77      & 0.776     & 0.782       & \textbf{0.809}       & 0.569        & 0.787        \\
watercraft & 0.513   & 0.632 & 0.591      & 0.355    & 0.471      & 0.579  &  0.607  & 0.601   & 0.582     & 0.594     & 0.583       & 0.603       & 0.467        & \textbf{0.608}        \\
\midrule
overall    & 0.560   & 0.596 & 0.635      & 0.352    & 0.552      & 0.626  & 0.659  & 0.642   & 0.634     & 0.661     & 0.645       & \textbf{0.670}       & 0.519        & 0.655       \\
    \midrule
    \textbf{F-Score@1\%} \\
    \midrule
airplane   & 0.412   & 0.487 & 0.446      & 0.415    & 0.376      & 0.494  & \textbf{0.598}  & 0.489   & -           & -          & 0.493       & 0.583       & 0.349        & 0.501        \\
bench      & 0.345   & 0.364 & 0.424      & 0.439    & 0.313      & 0.318  & 0.361  & 0.406   & -           & -           & 0.399       & \textbf{0.478}       & 0.229        & 0.459        \\
cabinet    & 0.327   & 0.316 & 0.381      & 0.350    & \textbf{0.450}      & 0.449  & 0.345  & 0.367   & -           & -           & 0.363       & 0.408       & 0.254        & 0.391        \\
car        & 0.481   & 0.514 & 0.481      & 0.319    & 0.486      & 0.315  & 0.304  & 0.497   & -           & -           & 0.523       & \textbf{0.564}       & 0.364        & 0.525        \\
chair      & 0.238   & 0.226 & 0.302      & 0.406    & 0.386      & 0.365  & \textbf{0.442}  & 0.334   & -           & -           & 0.262       & 0.309       & 0.193        & 0.269        \\
display    & 0.227   & 0.215 & 0.400      & 0.451    & 0.319      & \textbf{0.468}  & 0.466  & 0.310   & -           & -           & 0.253       & 0.296       & 0.187        & 0.274        \\
lamp       & 0.267   & 0.249 & 0.276      & 0.217    & 0.219      & 0.361  & \textbf{0.371}  & 0.315   & -           & -           & 0.287       & 0.315       & 0.289        & 0.307        \\
speaker    & 0.231   & 0.225 & \textbf{0.279}      & 0.199    & 0.190      & 0.249  & 0.200  & 0.211   & -           & -           & 0.256       & 0.152       & 0.218        & 0.278        \\
rifle       & 0.521   & 0.541 & 0.514      & 0.405    & 0.340      & 0.219  & 0.407  & 0.524   & -           & -           & 0.553       & \textbf{0.574}       & 0.417        & 0.563        \\
sofa       & 0.274   & 0.290 & 0.326      & 0.337    & 0.343      & 0.324  & 0.354  & 0.334   & -           & -           & 0.320       & \textbf{0.377}       & 0.192        & 0.367        \\
table      & 0.340   & 0.352 & 0.374      & 0.373    & 0.502      & \textbf{0.549}  & 0.461  & 0.419   & -           & -           & 0.385       & 0.406       & 0.276        & 0.369        \\
telephone  & 0.504   & 0.528 & 0.598      & 0.545    & 0.485      & 0.273  & 0.423  & 0.469   & -          & -           & 0.588       & \textbf{0.633}       & 0.355        & 0.572        \\
watercraft & 0.305   & 0.328 & 0.360      & 0.296    & 0.266      & 0.347  & 0.369  & 0.315   & -           & -           & 0.346       & \textbf{0.390}       & 0.274        & 0.386        \\
\midrule
overall    & 0.351   & 0.368 & 0.391      & 0.362    & 0.398      & 0.393  & 0.405  & 0.395   & -           & -           & 0.394       & \textbf{0.436}       & 0.282        & 0.404        \\
    \bottomrule
\end{tabular}
\caption{Comparison of single-view reconstruction between models on ShapeNet at $32^3$ resolution. Mean IoU and F-Score@1\% reported for all categories.}
  \label{tab:comparison_single_view}
\end{sidewaystable*}

\section{Comparison to other decoding schemes - Qualitative Results}
\label{appendix:decoding_schemes_qualitative_comparison}

\begin{figure*}
  \centering
  \includegraphics[page=1]{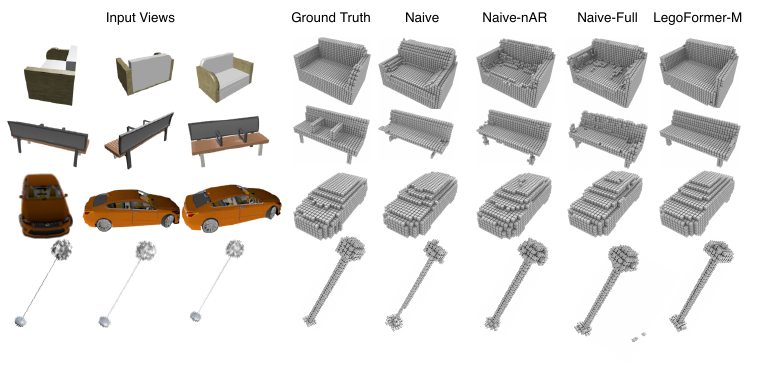}
  \caption{Multi-view object reconstructions of the four decoding schemes for 3 input views on ShapeNet with $32^3$ resolution. The second row shows an example of a challenging case for all decoding schemes.}
  \label{fig:scheme_comparison_qualitative}
\end{figure*}

In \cref{fig:scheme_comparison_qualitative} we show some qualitative examples of reconstructions using the different decoding schemes discussed in Section 3.2 of the main paper. In contrast to the naive approaches where the voxels are independently predicted, LegoFormer outputs are more structured and do not contain spike-like spurious full voxels (e.g., the small artifacts on the bottom right corner of the lamp reconstructed by Naive-Full). This observation supports our claim that using a tensor decomposition-based parametrization acts like a regularization and constrains the output space. The second row in the figure, is a relatively hard case for all approaches due to the unusual positioning of the armrests on the bench. While no model correctly reconstructs the 3D shape, all predictions are significantly noisier than the one obtained by LegoFormer-M. The same can be said of the reconstructed sofa in the first row, where all alternatives struggle to generate nice flat surfaces except for LegoFormer-M. 

\section{Reconstruction consistency on real-world data}
\label{appendix:incremental_real}

\begin{figure}
  \centering
  \includegraphics[page=1]{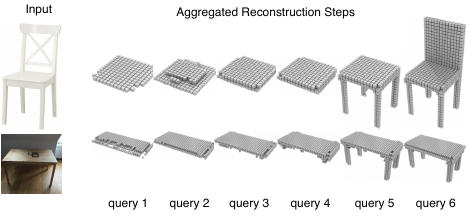}
  \caption{Incremental reconstruction on the Pix3D \cite{sun2018pix3d} dataset. For this experiment we used a LegoFormer-S trained using 6 decomposition queries to ease the visualization.}
  \label{fig:incremental_recons_real_imgs}
\end{figure}

We report in \cref{fig:incremental_recons_real_imgs} additional predictions on Pix3D \cite{sun2018pix3d} obtained by a LegoFormer-S model. Besides the full predicted voxel grids, we report also the individual outputs predicted by the decoder for each query incrementally aggregated together. This is similar to what we show in Fig. 7 of the main paper for LegoFormer-M. To ease the visualization of the individual part decomposition for this experiment we trained a variant of LegoFormer-S on ShapeNet\cite{chang2015shapenet} with only 6 output queries instead of the 12 used in the main paper.

The results confirms what highlighted for LegoFormer-M in the main paper: the queries tend to specialize into different type of 3D structures, with the first 4 mainly predicting flat horizontal planes and the last two focusing more on vertical surfaces like the chair and table legs (second to the last) or the backrest of the chair.
These results also show some limitations in the generalization performance of LegoFormer-S when applied to real data. For example the chair backrest is not properly reconstructed, likely because the color of the chair does not have much contrast with the white background and the network fails to detect it. 
We will discuss more about limitations in the following section. 

\section{Limitations}
\label{appendix:limitations}

\begin{figure}
  \centering
  \includegraphics[page=1]{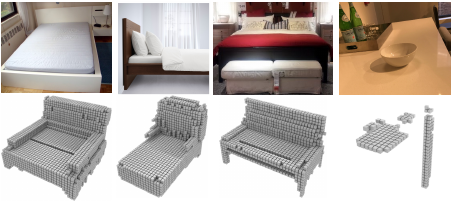}
  \caption{
 Failed reconstructions of LegoFormer-S on data from Pix3D \cite{sun2018pix3d}. We selected four challenging out-of-distribution samples as they belong to the \textit{bed} and \textit{bowl} categories which are not present in the training set. While for the beds the model tries to come up with predictions which are a mix of a sofa and an arm-chair (the two most related classes in the training set), for the bowl the model fails completely.
}
  \label{fig:failure_unseen}
\end{figure}

We show here some limitations of LegoFormer-S when applied to the challenging real data from Pix3D\cite{sun2018pix3d}. For all the following results we use LegoFormer-S trained on the ShapeNet\cite{chang2015shapenet} dataset and tested without fine tuning it on the real images. 
This setting is particularly challenging due to the domain gap existing between real and synthetic data and helps to highlight failures of the network.   
During our experiments we identified several sources of "failure":

\textbf{Unseen Categories.} In \cref{fig:failure_unseen} we show four examples of failed reconstructions of objects outside the categories available in the training set. The first three images represent \textit{close} out-of-distribution samples belonging to the  \textit{bed} category which is not present in ShapeNet but it is relatively similar to other furniture categories. For this reason the network generates a 3D model that resembles a mix between a sofa and an arm chair, two of the available categories in the training set. While wrong in terms of overall shape and proportions, the predictions do try to capture some of the characteristics of the images, like the elongated shape of the bed in the second column. The last column shows a \text{far} out-of-distribution sample belonging to the \textit{bowl} category. There isn't a similar category in the ShapeNet training data and for this reason the model fails generating a meaningless output. We believe that these limitations could be alleviated by training on a more varied synthetic dataset to enlarge the set of categories considered in-distribution.

\begin{figure}
  \centering
  \includegraphics[page=1]{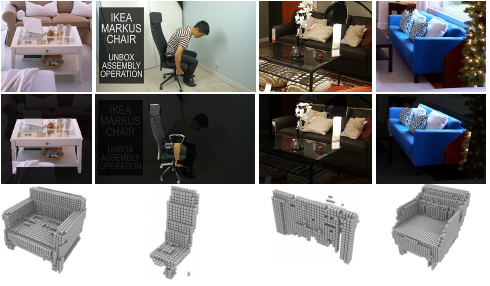}
  \caption{
    Failed reconstructions of LegoFormer-S on data from Pix3D \cite{sun2018pix3d} in presence of occlusions. The first row shows the original image, the second row the image segmented according to the GT object segmentation mask (the input to LegoFormer-S), finally the last row show the 3D model predicted. Due to occlusions our model is mislead into predicting objects from different categories (column one), can generate only partial shape (column two) or fails the reconstruction altogether due to too much clutter (column three) or misleading perspective (column four).
}
  \label{fig:failure_occlusion}
\end{figure}

\textbf{Occlusions.} While ShapeNet \cite{chang2015shapenet} always depicts objects perfectly visible and without any kind of occlusions, this is not true for the real data. 
For this reason the model might generate wrong predictions in such cases. We report some of these examples in \cref{fig:failure_occlusion}. 
In the first column the network is probably confused by the presence of objects on the table surface and reconstructs the table as a chair. In this case occlusions cause a non uniform "texture" of the object, all ShapeNet \cite{chang2015shapenet} models have instead uniform textures, however sofas and armchairs can exhibit non uniform colors due to shading effects.
The second column shows an example of partial occlusions since the arms of the chair are occluded by the man sitting on it. In this case the network reconstructs only the clearly visible part of the object resulting in an arm-less chair.
Finally the third and fourth columns show additional examples where distracting elements (the pillows) and occlusions makes once again the model fail badly (third column) or be mislead on the proportion of the object (fourth column). 
We believe that this weakness could be partially overcome by simulating occlusions during training, which we are not currently doing. Moreover we show here results for the single-view model, but part of these failures could maybe be overcome in the presence of multiple views, where the occluded parts should change from view to view.


\begin{figure}
  \centering
  \vspace{-2cm}
  \includegraphics[width=\columnwidth]{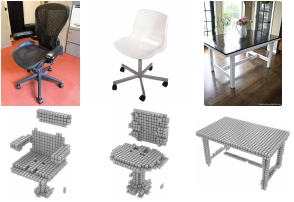}
  \caption{
  Failure cases for LegoFormer-S on Pix3D \cite{sun2018pix3d} data due to failed rank-1 tensor decomposition for the output grid. The model fails to predict tiny structures like the chair and table legs.
}
  \label{fig:failure_tensor_decomp}
\end{figure}

\textbf{Failed tensor decomposition.} In \cref{fig:failure_tensor_decomp} we show three examples where our proposed tensor decomposition fails to generate some details of the real models. For example the legs of the chairs and the bottom part of the table are missing or broken in the reconstruction. As any other compression method a rank-1 decomposition can have the side effect of removing details. 
We believe that part of the missing details, if needed, could be recovered from the raw output of LegoFormer pre-thresholding with an additional small refiner network similar to what done in \cite{xie2020pix2vox++}.


\begin{figure}
  \centering
  \vspace{-10cm}
  \includegraphics[page=1]{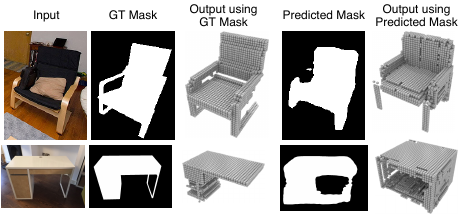}
  \caption{Model reconstructed by a LegoFormer-S model on Pix3D \cite{sun2018pix3d} using GT masks to segment the object from the background (third column) or predicted masks (last column). The use of non-optimal masks lead to a clear degradation in performance.}
  \label{fig:mask_comparison}
\end{figure}

\textbf{Bad Segmentation Mask.} All the results so far have used the GT segmentation masks to segment out the object from the background, in \cref{fig:mask_comparison} we show the impact on performance of using predicted masks from a Mask-RCNN model \cite{he2017mask} trained on COCO \cite{lin2014microsoft}. 
As expected non perfect masks lead to a degradation on the quality of the predicted models. Fine-tuning the Mask-RCNN model on the desired categories can certainly help to alleviate these issues by improving the segmentation performance.
Another option could be to train our model using synthetic rendering placed in front of random background. If this training succeeds we could potentially drop the dependency from a nicely segmented input altogether.


\end{appendices}

\end{document}